\documentclass[runningheads]{llncs}
\usepackage{graphicx}
\usepackage{amsmath,amssymb} 
\usepackage{color}
\usepackage[width=122mm,left=12mm,paperwidth=146mm,height=193mm,top=12mm,paperheight=217mm]{geometry}
\usepackage{url}

\begin{document}
\newcommand{\todo}[1]{{\color{red} {\bf TODO:} \it #1}}
\newcommand{\ning}[1]{{\color{blue}{\bf Ning:} \it #1}}
\renewcommand{\tabcolsep}{0.2cm}

\pagestyle{headings}
\mainmatter

\title{Part-based R-CNNs for \\ Fine-grained Category Detection}

\titlerunning{Part-based R-CNNs for Fine-grained Category Detection}

\authorrunning{Zhang, Donahue, Girshick, Darrell}

\author{Ning Zhang, Jeff Donahue, Ross Girshick, Trevor Darrell\\
\texttt{\{nzhang,jdonahue,rbg,trevor\}@eecs.berkeley.edu}}
\institute{University of California, Berkeley}

\maketitle

\begin{abstract}

Semantic part localization can facilitate fine-grained categorization by explicitly isolating subtle appearance differences associated with specific object parts.  Methods for pose-normalized representations have been proposed, but generally presume bounding box annotations at test time due to the difficulty of object detection. We propose a model for fine-grained categorization that overcomes these limitations by leveraging deep convolutional features computed on bottom-up region proposals. Our method learns whole-object and part detectors, enforces learned geometric constraints between them, and predicts a fine-grained category from a pose-normalized representation.
Experiments on the Caltech-UCSD bird dataset confirm that our method outperforms state-of-the-art fine-grained categorization methods in an end-to-end evaluation without requiring a bounding box at test time.

\keywords{Fine-grained recognition, object detection, convolutional models}
\end{abstract}

\section{Introduction}

\begin{figure}[h]
\begin{center}
   \includegraphics[width=0.99\linewidth]{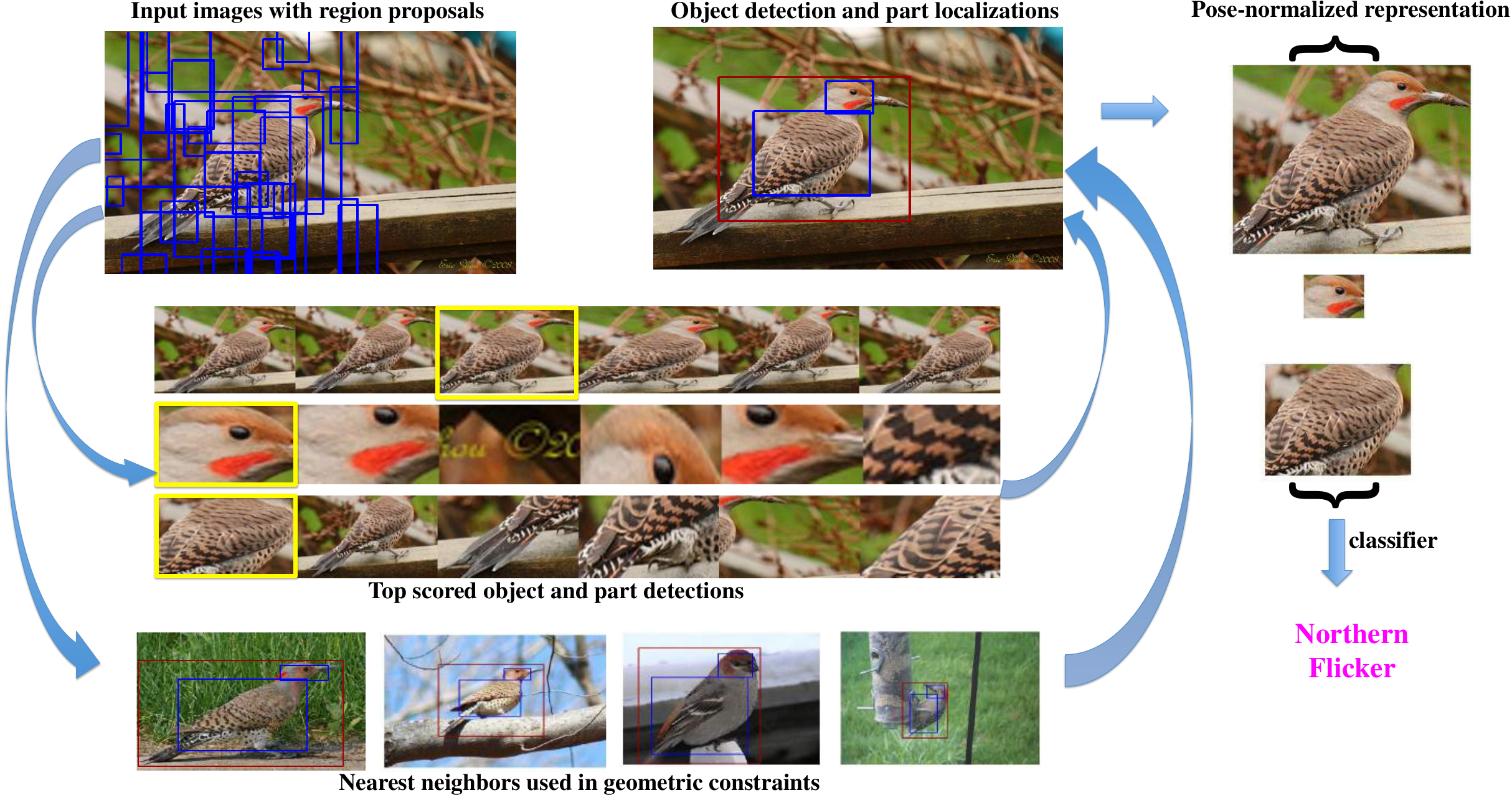}
\end{center}
   \caption{\textbf{Overview of our part localization}  Starting from bottom-up region proposals (top-left), we train both object and part detectors based on deep convolutional features. During test time, all the windows are scored by all detectors (middle), and we apply non-parametric geometric constraints (bottom) to rescore the windows and choose the best object and part detections (top-right).  The final step is to extract features on the localized semantic parts for fine-grained recognition for a pose-normalized representation and then train a classifier for the final categorization. Best viewed in color.} 
\label{fig:concept}
\end{figure}

The problem of visual fine-grained categorization can be extremely challenging due to the subtle differences in the appearance of certain parts across related categories. In contrast to basic-level recognition, fine-grained categorization aims to distinguish between different breeds or species or product models, and often requires distinctions that must be conditioned on the object pose for reliable identification. Facial recognition is the classic case of fine-grained recognition, and it is noteworthy that the best facial recognition methods jointly discover facial landmarks and extract features from those locations. 

Localizing the parts in an object is therefore central to establishing correspondence between object instances and discounting object pose variations and camera view position. Previous work has investigated part-based approaches to this problem~\cite{poof,BirdletsFarrellICCV11,iccv13_keypoint,UW_NIPS12,PosePoolingKernelsZhangEtalCVPR12,Goering14:NPT}. The bottleneck for many pose-normalized representations
is indeed accurate part localization.
The Poselet \cite{BourdevMalikICCV09} and DPM \cite{dpm} methods have previously been utilized to obtain part localizations with a modest degree of success; methods generally report adequate part localization only when given a known bounding box at test time~\cite{Tricos_Chai_ECCV12,iccv13_alignment,ParkhiEtalICCV11,ParkhiEtalCVPR12,iccv13_partmatching}. By developing a novel deep part detection scheme, we propose an end-to-end fine grained categorization system which requires no knowledge of object bounding box at test time, and can achieve performance rivaling previously reported methods requiring the ground truth bounding box at test time to filter false positive detections.

The recent success of convolutional networks, like~\cite{krizhevsky}, on the ImageNet Challenge~\cite{ILSVRC} has inspired further work on applying deep convolutional features to related image classification~\cite{decaf} and detection tasks~\cite{rcnn}.
In~\cite{rcnn}, Girshick et al. achieved breakthrough performance on object detection by applying the CNN of~\cite{krizhevsky} to a set of bottom-up candidate region proposals~\cite{selsearch}, boosting PASCAL detection performance by over 30\% compared to the previous best methods.
Independently, OverFeat~\cite{overfeat} proposed localization using a CNN to regress to object locations.
However, the progress of leveraging deep convolutional features is not limited to basic-level object detection.
In many applications such as fine-grained recognition, attribute recognition, pose estimation, and others, reasonable predictions demand accurate part localization.

Feature learning has been used for fine-grained recognition and attribute estimation, but was limited to engineered features for localization.
DPD-DeCAF~\cite{dpd} used DeCAF~\cite{decaf} as a feature descriptor, but relied on HOG-based DPM~\cite{dpm} for part localization.
PANDA~\cite{panda} learned part-specific deep convolutional networks whose location was conditioned on HOG-based poselet models.
These models lack the strength and detection robustness of R-CNN~\cite{rcnn}.
In this work we explore a unified method that uses the same deep convolutional representation for detection as well as part description.

We conjecture that progress made on bottom-up region proposal methods, like selective search~\cite{selsearch}, could benefit localization of smaller parts in addition to whole objects.
As we show later, average recall of parts using selective search proposals is 95\% on the Caltech-UCSD bird dataset.

In this paper, we propose a part localization model which overcomes the limitations of previous fine-grained recognition systems by leveraging deep convolutional features computed on bottom-up region proposals.
Our method learns part appearance models and enforces geometric constraints between parts.
An overview of our method is shown in Figure~\ref{fig:concept}. We have investigated different geometric constraints, including a non-parametric model of joint part locations conditioned on nearest neighbors in semantic appearance space.
We present state-of-the-art results evaluating our approach on the widely used fine-grained benchmark Caltech-UCSD bird dataset~\cite{DatasetCUB200}.

%
%
%

\section{Related work}

\subsection{Part-based models for detection and pose localization}
Previous work has proposed explicit modeling of object part appearances and locations for more accurate recognition and localization.
Starting with pictorial structures~\cite{pedro2000,pictorial},
and continuing through poselets~\cite{BourdevMalikICCV09} and related work, many methods have jointly localized a set of geometrically related parts.
The deformable parts model (DPM)~\cite{dpm}, until recently the state-of-the-art PASCAL object detection method, models parts with additional learned filters in positions anchored with respect to the whole object bounding box, allowing parts to be displaced from this anchor with learned deformation costs.
The ``strong'' DPM~\cite{Hossein_ECCV12} adapted this method for the strongly supervised setting in which part locations are annotated at training time.
A limitation of these methods is their use of weak features (usually HOG~\cite{hog}). 

\subsection{Fine-grained categorization}
Recently, a large body of computer vision research has focused on the fine-grained classification problem
in a number of domains, such as animal breeds or species~\cite{BirdletsFarrellICCV11,KhoslaYaoJayadevaprakashFeiFei_FGVC2011,Liu_Dogs_2012,MartinezMunozEtalCVPR2009,ParkhiEtalCVPR12,YaoEtalCVPR11}, plant species~\cite{AngelovaCVPR13,Anelia13,leafsnap,nilsback_visual_2006,nilsback_automated_2008,VantageFramesCVPR13}, and man-made objects~\cite{maji13fine-grained,StarkKrauseBMVC12}. 

Several approaches are based on detecting and extracting features from certain parts of objects. Farrell et al.~\cite{BirdletsFarrellICCV11} proposed a pose-normalized representation using poselets \cite{BourdevMalikICCV09}. Deformable part models \cite{dpm} were used in \cite{ParkhiEtalCVPR12,dpd} for part localization. Based on the work of localizing fiducial landmarks on faces~\cite{Belhumeur_Localizing_2011}, Liu et al.~\cite{Liu_Dogs_2012} proposed an exemplar-based geometric method to detect dog faces and extract highly localized image features from keypoints to differentiate dog breeds. Furthermore, Berg et al.~\cite{poof} learned a set of highly discriminative intermediate features by learning a descriptor for each pair of keypoints. Moreover, in \cite{iccv13_keypoint}, the authors extend the non-parametric exemplar-based method of \cite{Belhumeur_Localizing_2011} by enforcing pose and subcategory consistency. Yao et al.~\cite{YaoEtalCVPR12} and Yang et al.~\cite{UW_NIPS12} have investigated template matching methods to reduce the cost of sliding window approaches. Recent work by G\"{o}ring et al.~\cite{Goering14:NPT} transfers part annotations from objects with similar global shape as non-parametric part detections. 
All these part-based methods, however, require the ground truth bounding box at test time for part localization or keypoint prediction. 

Human-in-the-loop methods \cite{BransonEtalECCV10,Deng_FineCrowd_2013,DuanEtalCVPR12} ask a human to name attributes of the object, click on certain parts or mark the most discriminative regions to improve classification accuracy.
Segmentation-based approaches are also very effective for fine-grained recognition.
Approaches such as \cite{Tricos_Chai_ECCV12,iccv13_alignment,ParkhiEtalICCV11,ParkhiEtalCVPR12,iccv13_partmatching} used region-level cues to infer the foreground segmentation mask and to discard the noisy visual information in the background.
Chai et al.~\cite{iccv13_symbiotic} showed that jointly learning part localization and foreground segmentation together can be beneficial for fine-grained categorization.
Similar to most previous part-based approaches, these efforts require the ground truth bounding box to initialize the segmentation seed.
In contrast, the aim of our work is to perform end-to-end fine-grained categorization with no knowledge at test time of the ground truth bounding box.
Our part detectors use convolutional features on bottom-up region proposals, together with learned non-parametric geometric constraints to more accurately localize object parts, thus enabling strong fine-grained categorization.




\subsection{Convolutional networks}
In recent years, convolutional neural networks (CNNs) have been incorporated into a number of visual recognition systems in a wide variety of domains.
At least some of the strength of these models lies in their ability to \textit{learn} discriminative features from raw data inputs (e.g., image pixels), in contrast to more traditional object recognition pipelines which compute hand-engineered features on images as an initial preprocessing step.
CNNs were popularized by LeCun and colleagues who initially applied such models to digit recognition~\cite{Lecun89} and OCR~\cite{Lecun98OCR} and later to generic object recognition tasks~\cite{jarrett-iccv2009}.
With the introduction of large labeled image databases~\cite{ILSVRC} and GPU implementations used to efficiently perform the massive parallel computations required for learning and inference in large CNNs,
these networks have become the most accurate method for generic object classification~\cite{krizhevsky}.

Most recently, generic object detection methods have begun to leverage deep CNNs and outperformed any competing approaches based on traditional features.
OverFeat~\cite{overfeat} uses a CNN to regress to object locations in a coarse sliding-window detection framework.
Of particular inspiration to our work is the R-CNN method~\cite{rcnn} which leverages features from a deep CNN in a region proposal framework to achieve unprecedented object detection results on the PASCAL VOC dataset.
Our method generalizes R-CNN by applying it to model object parts in addition to whole objects, which our empirical results will demonstrate is essential for accurate fine-grained recognition.


\section{Part-based R-CNNs}
\label{sec:recursive-part-dets}

While~\cite{rcnn} demonstrated the effectiveness of the R-CNN method on a generic object detection task (PASCAL VOC), it did not explore the application of this method to simultaneous localization and fine-grained recognition.
Because our work operates in this regime, we extend R-CNN to detect objects and localize their parts under a geometric prior.
With hypotheses for the locations of individual semantic parts of the object of interest (e.g., the location of the head for an animal class), it becomes reasonable to model subtle appearance differences which tend to appear in locations that are roughly fixed with respect to these parts.

In the R-CNN method, for a particular object category, a candidate detection $x$ with CNN feature descriptor $\phi(x)$ is assigned a score of $w_0^{\intercal} \phi(x)$, where $w_0$ is the learned vector of SVM weights for the object category.
In our method, we assume a strongly supervised setting (e.g., \cite{Hossein_ECCV12}) in which at training time we have ground truth bounding box annotations not only for full objects, but for a fixed set of semantic parts $\{p_1, p_2, ..., p_n\}$ as well.

Given these part annotations, at training time all objects and each of their parts are initially treated as independent object categories: we train a one-versus-all linear SVM on feature descriptors extracted over region proposals, where regions with $\ge 0.7$ overlap with a ground truth object or part bounding box are labeled as positives for that object or part, and regions with $\le 0.3$ overlap with any ground truth region are labeled as negatives.
Hence for a single object category we learn whole-object (``root'') SVM weights $w_0$ and part SVM weights $\{w_1, w_2, ..., w_n\}$ for parts $\{p_1, p_2, ..., p_n\}$ respectively.
At test time, for each region proposal window we compute scores from all root and part SVMs.
Of course, these scores do not incorporate any knowledge of how objects and their parts are constrained geometrically; for example, without any additional constraints the \textit{bird head} detector may fire outside of a region where the \textit{bird} detector fires.
Hence our final joint object and part hypotheses are computed using the geometric scoring function detailed in the following section, which enforces the intuitively desirable property that pose predictions are consistent with the statistics of poses observed at training time.

\subsection{Geometric constraints}
Let $X =\{x_0, x_1, \ldots, x_n\}$ denote the locations (bounding boxes) of object $p_0$ and $n$ parts $\{p_i\}_{i=1}^n$, which are annotated in the training data, but unknown at test time. Our goal is to infer both the object location and part locations in a previously unseen test image. Given the R-CNN  weights $\{w_0, w_1, \ldots, w_n\}$ for object and parts, we will have the corresponding detectors $\{d_0, d_1, \ldots, d_n\}$ where each detector score is $d_i(x) = \sigma(w_i^{\intercal} \phi(x))$, where $\sigma(\cdot)$ is the sigmoid function and $\phi(x)$ is the CNN feature descriptor extracted at location $x$.
We infer the joint configuration of the object and parts by solving the following optimization problem:
\begin{equation}
X^* = \arg \max_{X}
\Delta(X)
\prod_{i=0}^n d_i(x_i)
\end{equation}
where $\Delta(X)$ defines a scoring function over the joint configuration of the object and root bounding box.
We consider and report quantitative results on several configuration scoring functions $\Delta$, detailed in the following paragraphs.


\paragraph{Box constraints.}
One intuitive idea to localize both the object and parts is to consider each possible object window and all the windows inside the object and pick the windows with the highest part scores.
In this case, we define the scoring function
\begin{equation}
\Delta_{\mathrm{box}}(X) = \prod_{i=1}^n c_{x_0}(x_i)
\end{equation}
where
\begin{equation}
c_x(y) =
\left\{
\begin{array}{ll}
1 & \text{if region } y \text{ falls outside region } x \text{ by at most } \epsilon \text{ pixels }\\
0 & \text{otherwise}
\end{array}
\right.
\end{equation}
In our experiments, we let $\epsilon = 10$.

\begin{figure*}
\begin{center}
\includegraphics[width=0.15\linewidth, height = 0.1\linewidth]{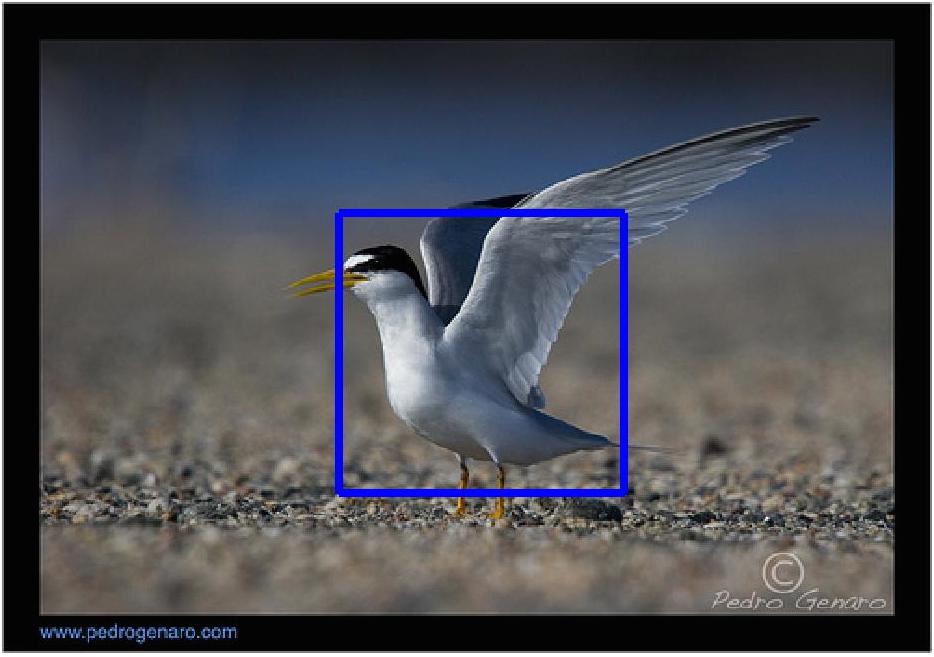} 
\includegraphics[width=0.15\linewidth, height = 0.1\linewidth]{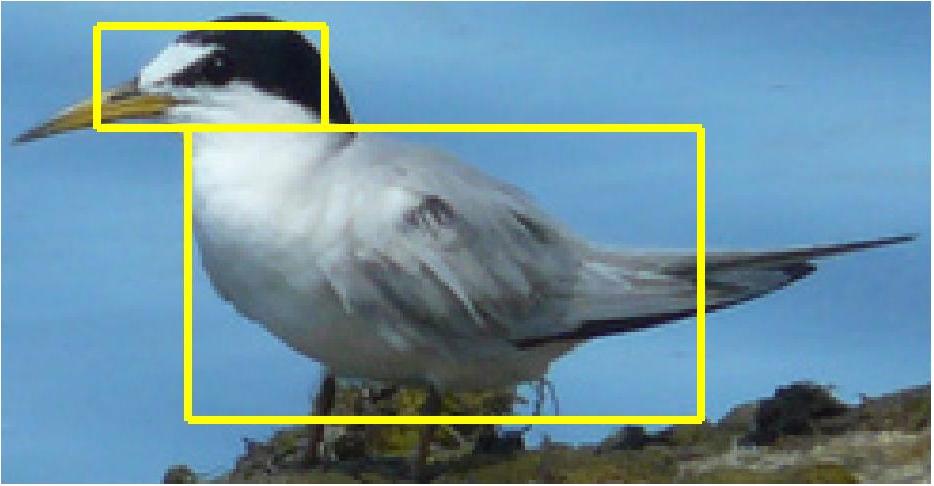} 
\includegraphics[width=0.15\linewidth, height = 0.1\linewidth]{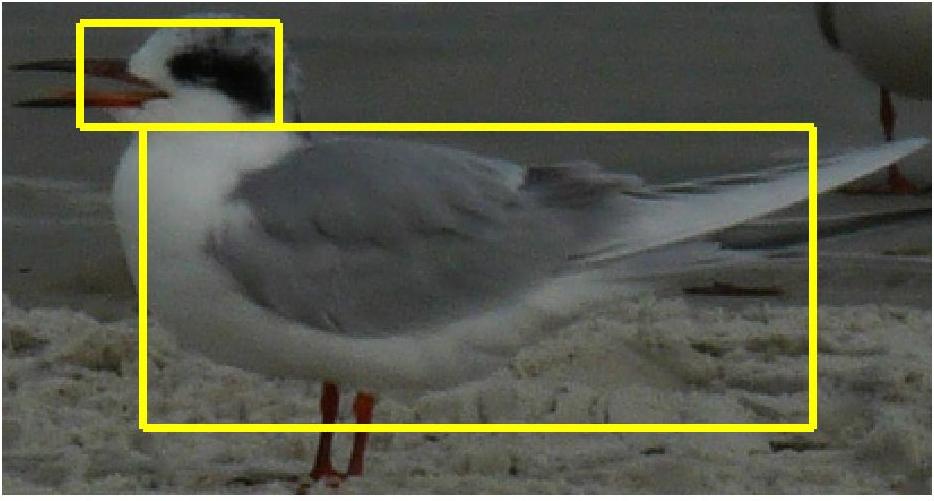} 
\includegraphics[width=0.15\linewidth, height = 0.1\linewidth]{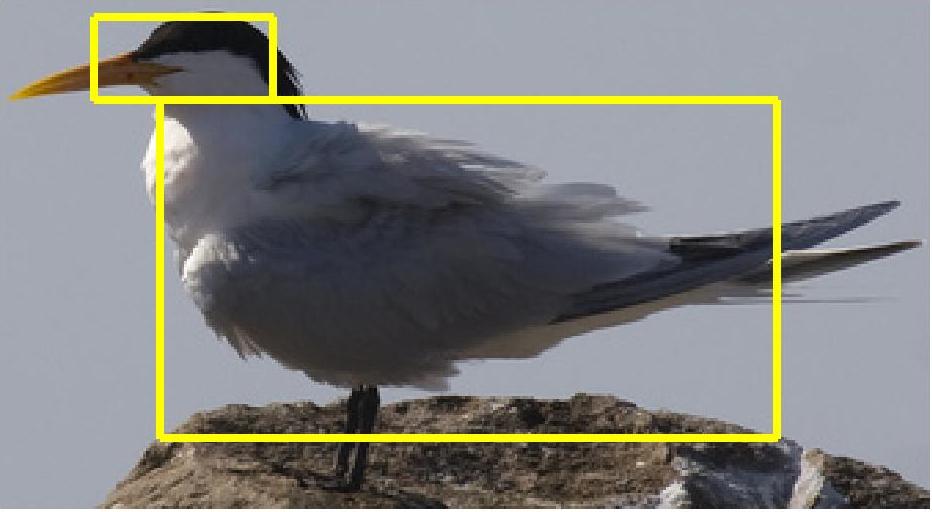} 
\includegraphics[width=0.15\linewidth, height = 0.1\linewidth]{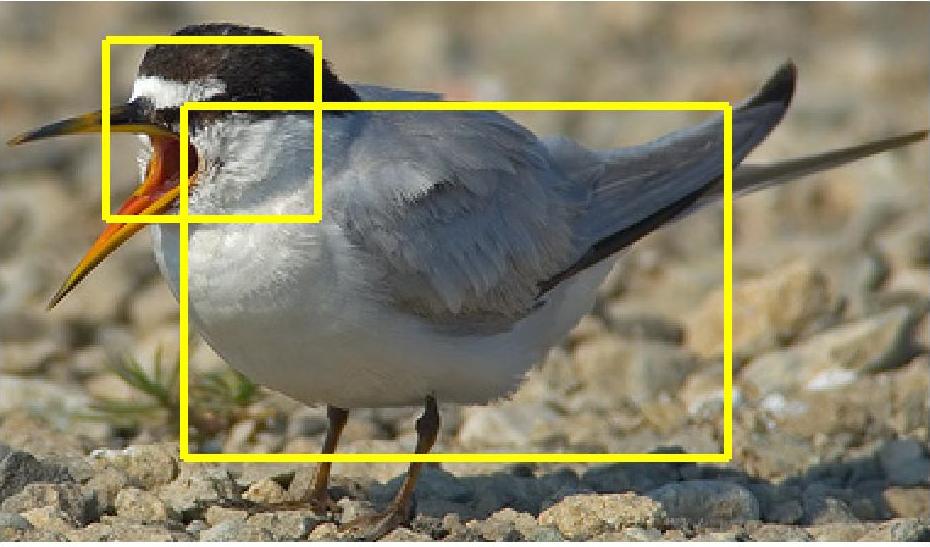} 
\includegraphics[width=0.15\linewidth, height = 0.1\linewidth]{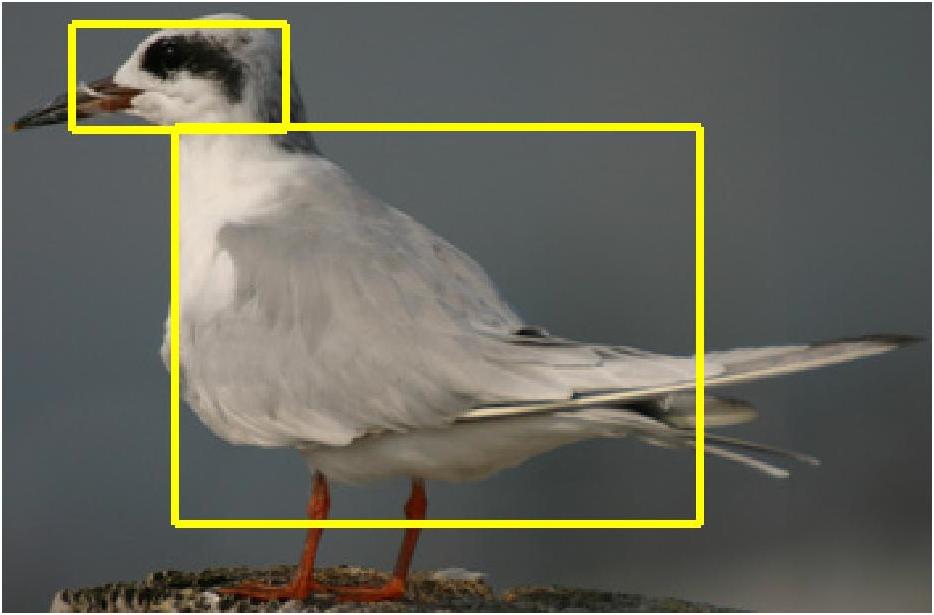} 
\\
\includegraphics[width=0.15\linewidth, height = 0.1\linewidth]{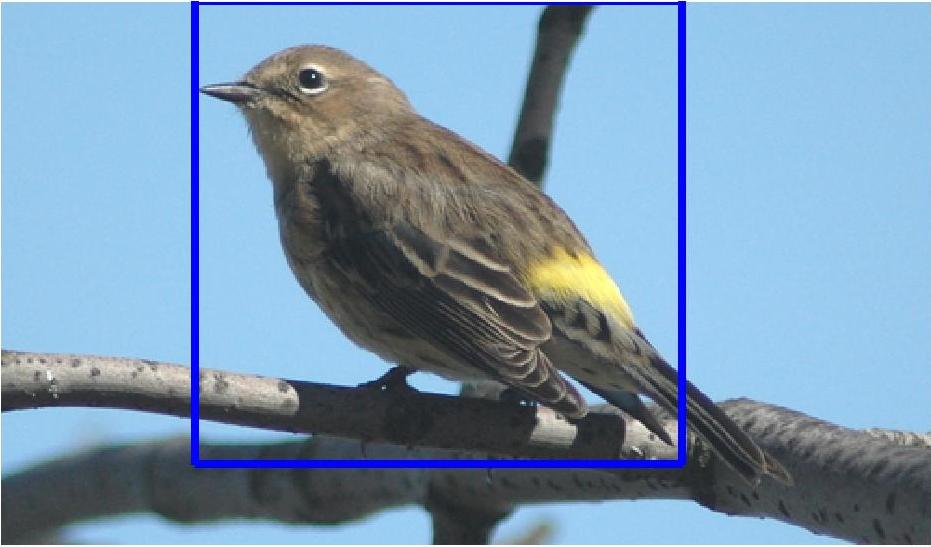} 
\includegraphics[width=0.15\linewidth, height = 0.1\linewidth]{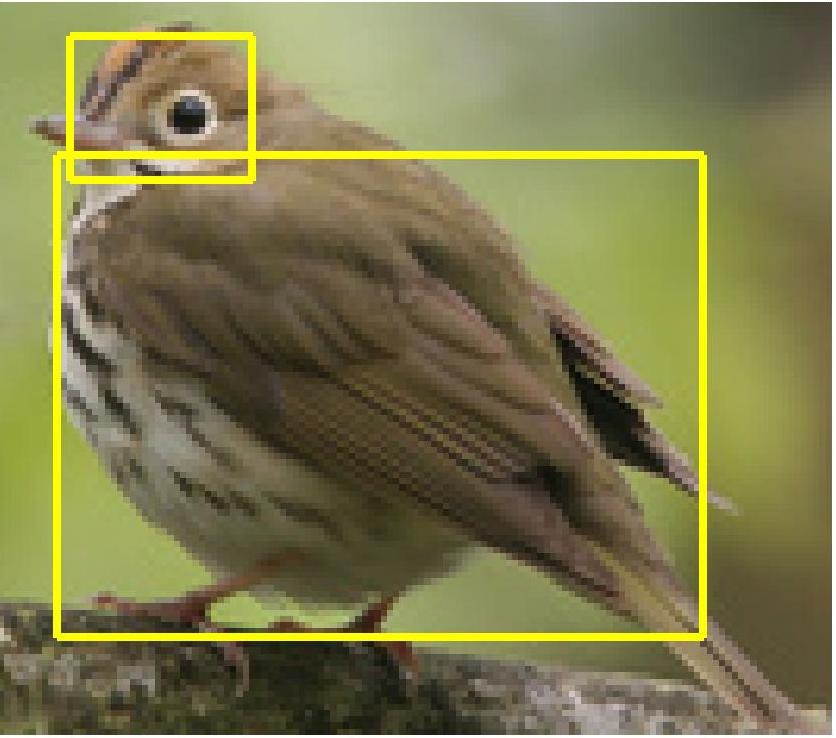} 
\includegraphics[width=0.15\linewidth, height = 0.1\linewidth]{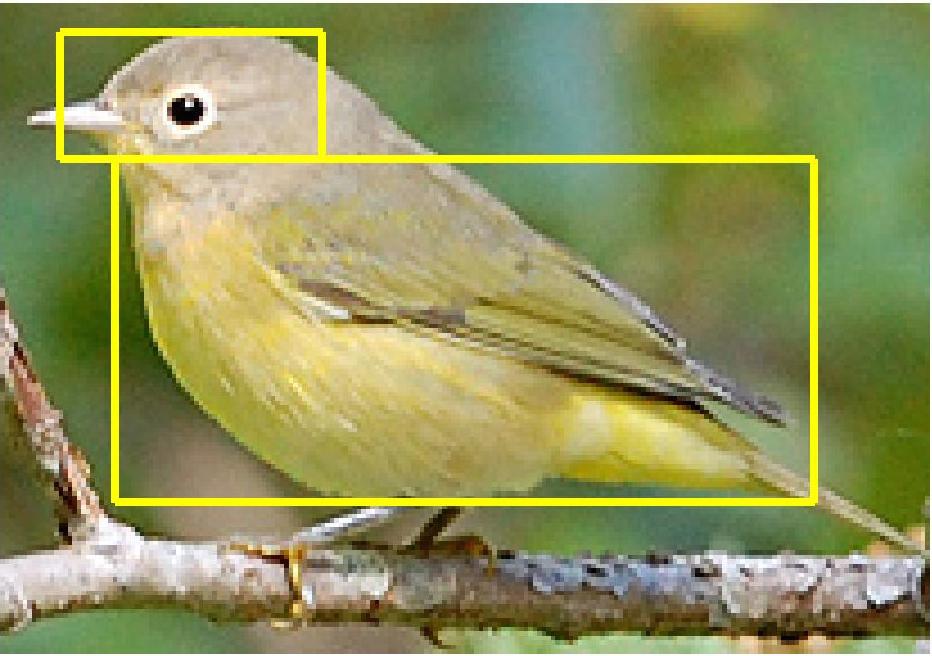} 
\includegraphics[width=0.15\linewidth, height = 0.1\linewidth]{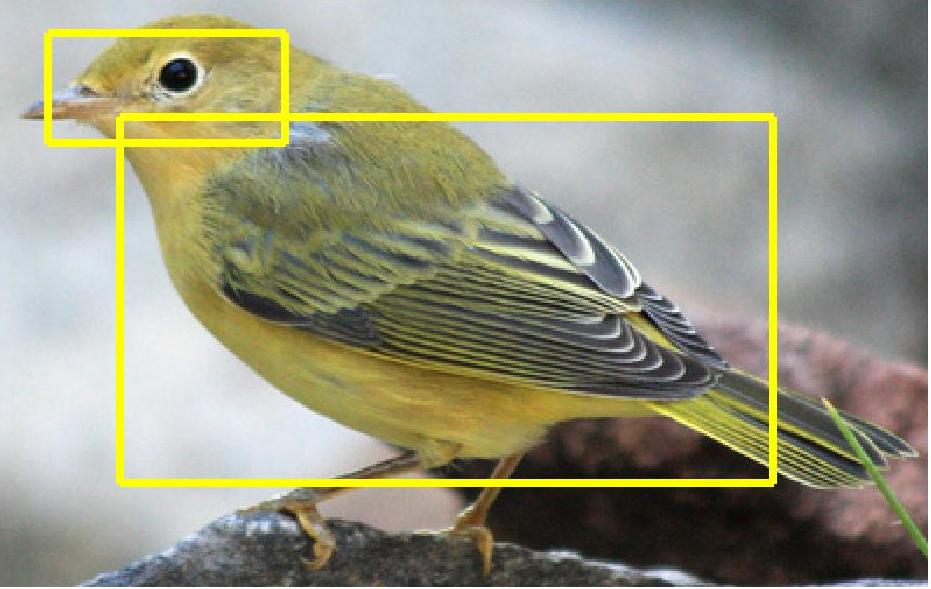} 
\includegraphics[width=0.15\linewidth, height = 0.1\linewidth]{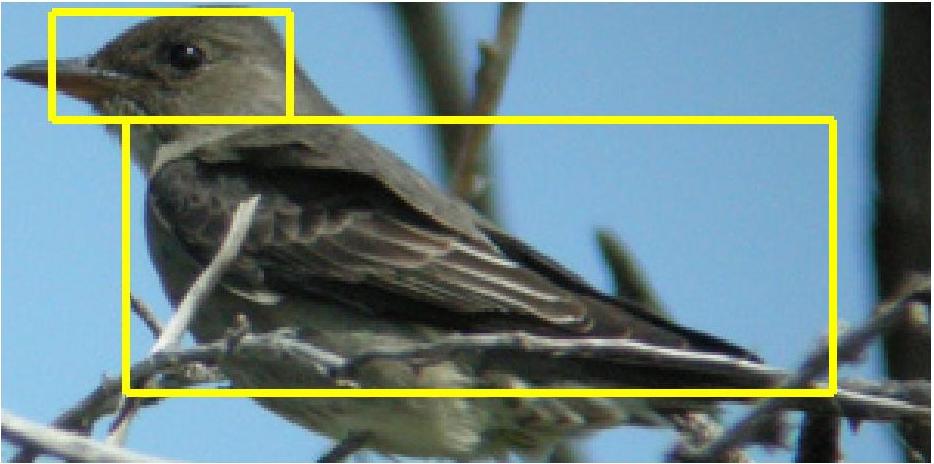} 
\includegraphics[width=0.15\linewidth, height = 0.1\linewidth]{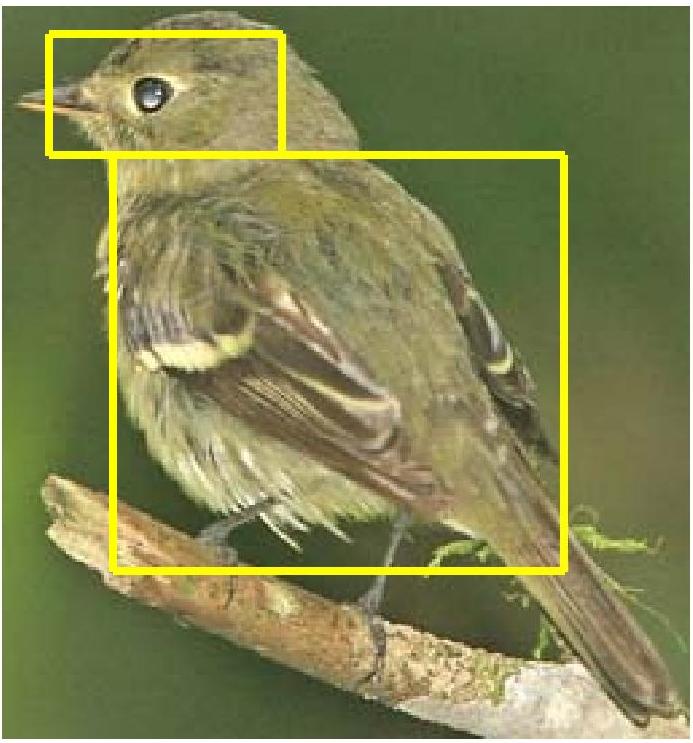} 
\\
\includegraphics[width=0.15\linewidth, height = 0.1\linewidth]{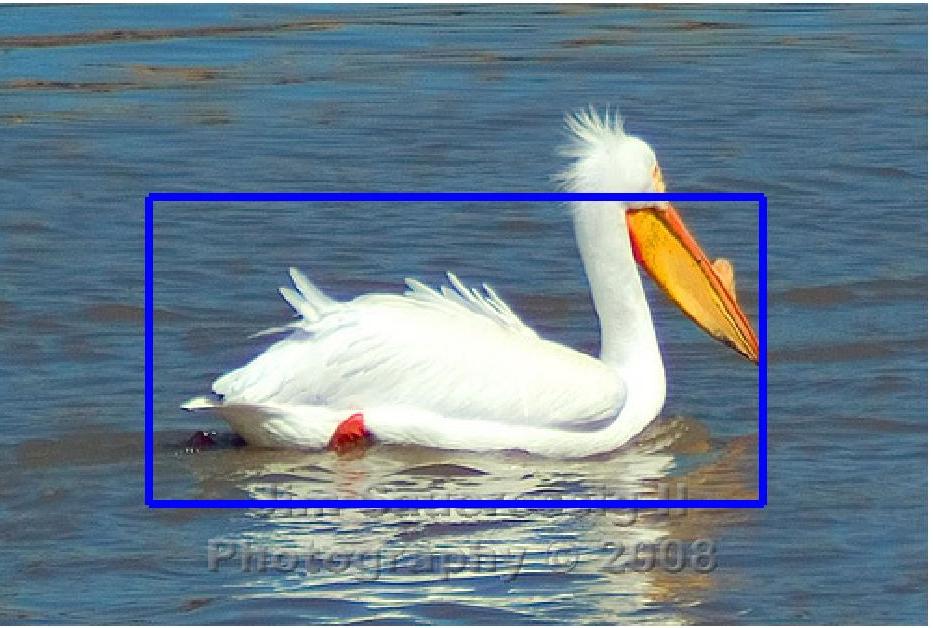} 
\includegraphics[width=0.15\linewidth, height = 0.1\linewidth]{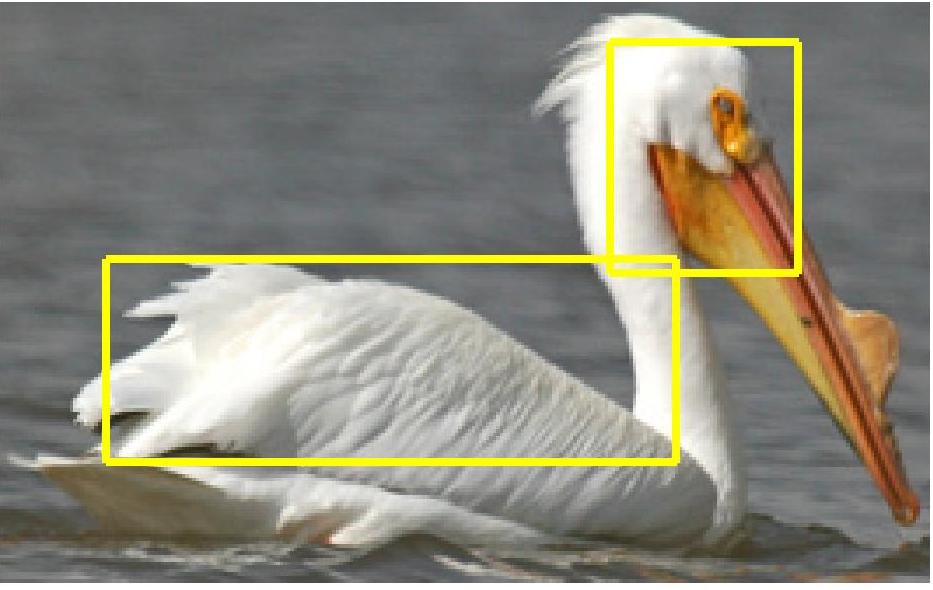} 
\includegraphics[width=0.15\linewidth, height = 0.1\linewidth]{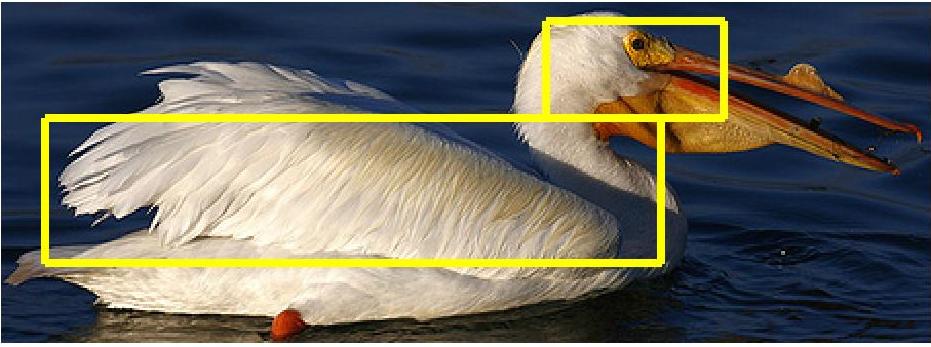} 
\includegraphics[width=0.15\linewidth, height = 0.1\linewidth]{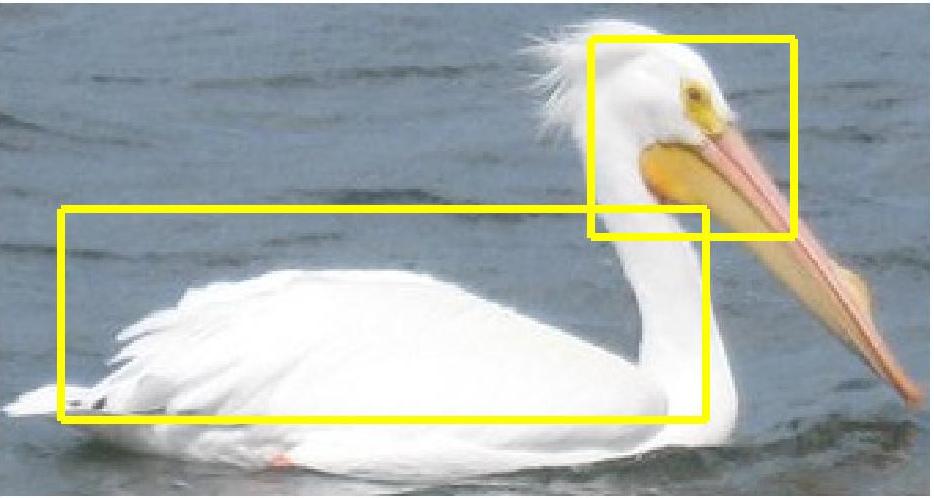} 
\includegraphics[width=0.15\linewidth, height = 0.1\linewidth]{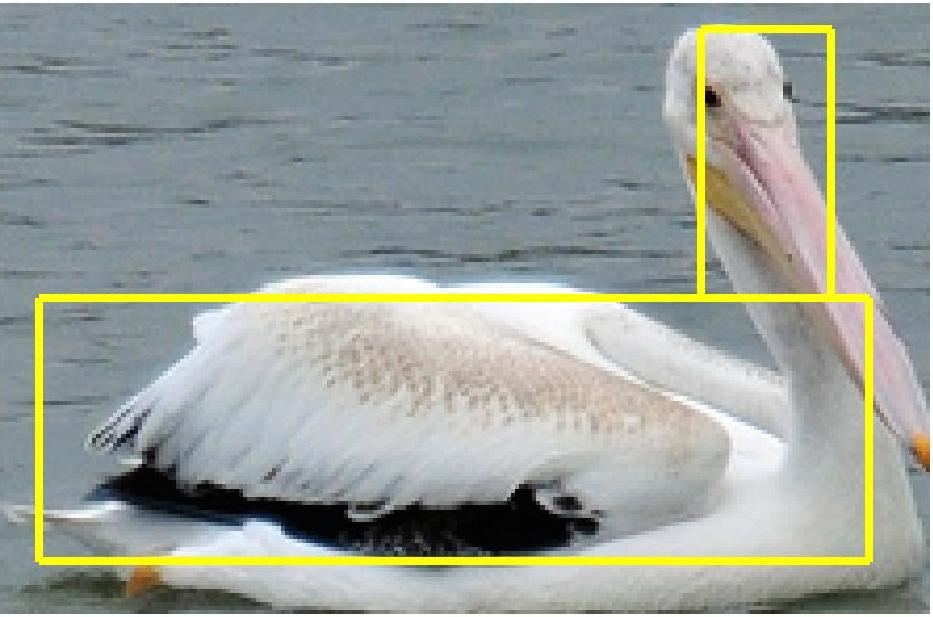} 
\includegraphics[width=0.15\linewidth, height = 0.1\linewidth]{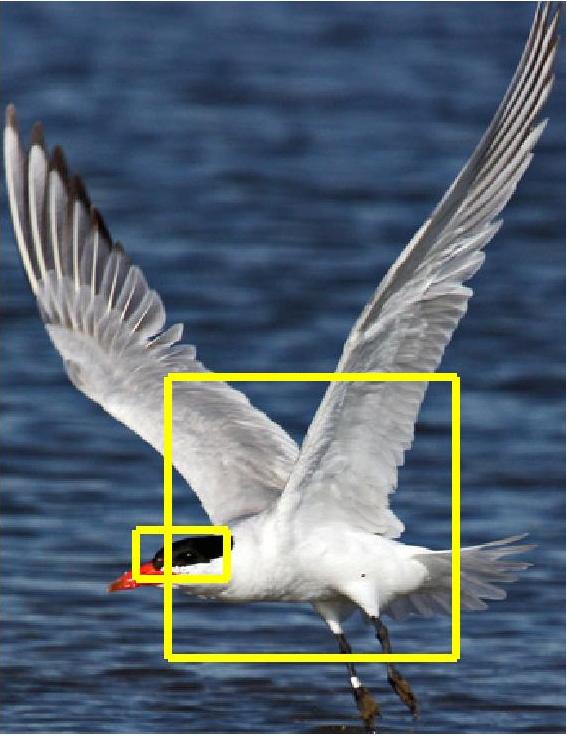} 
\\
\includegraphics[width=0.15\linewidth, height = 0.1\linewidth]{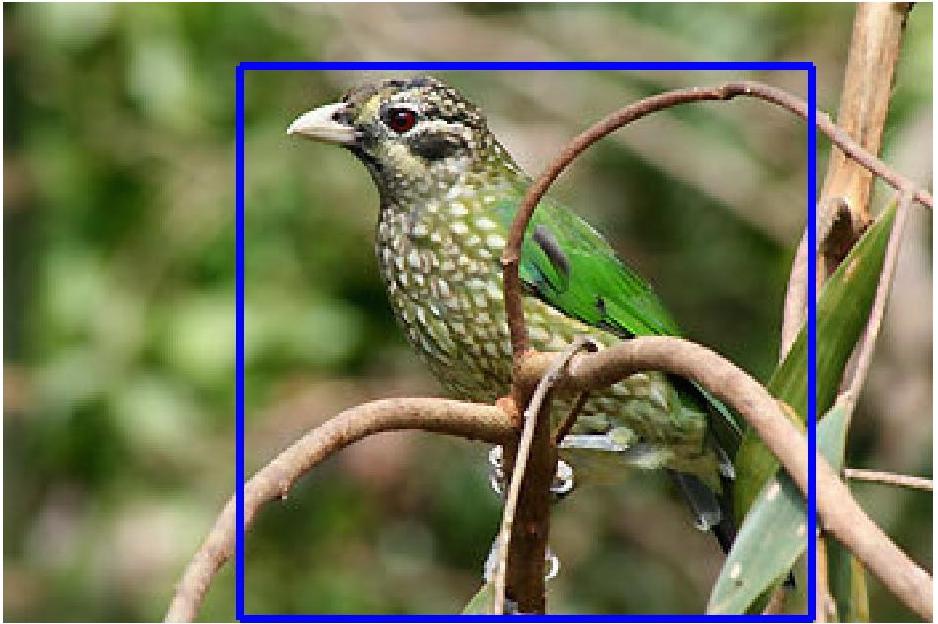} 
\includegraphics[width=0.15\linewidth, height = 0.1\linewidth]{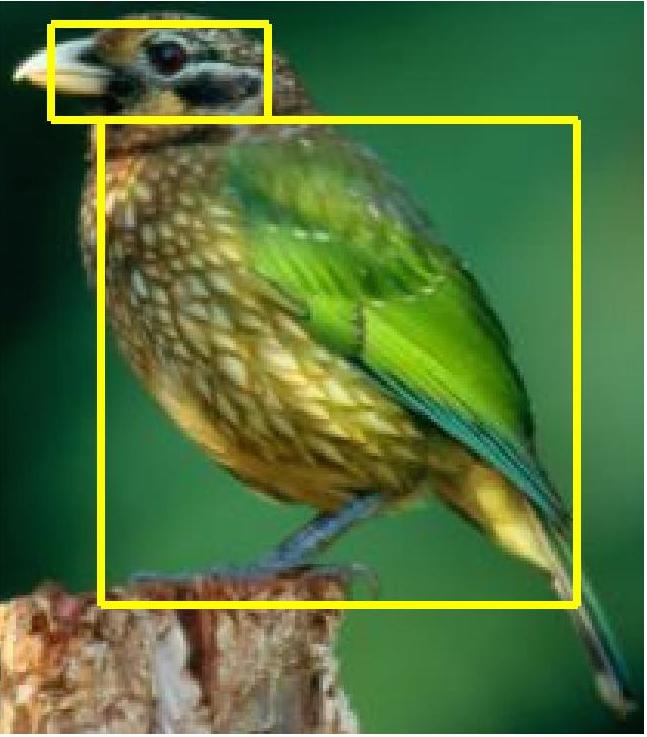} 
\includegraphics[width=0.15\linewidth, height = 0.1\linewidth]{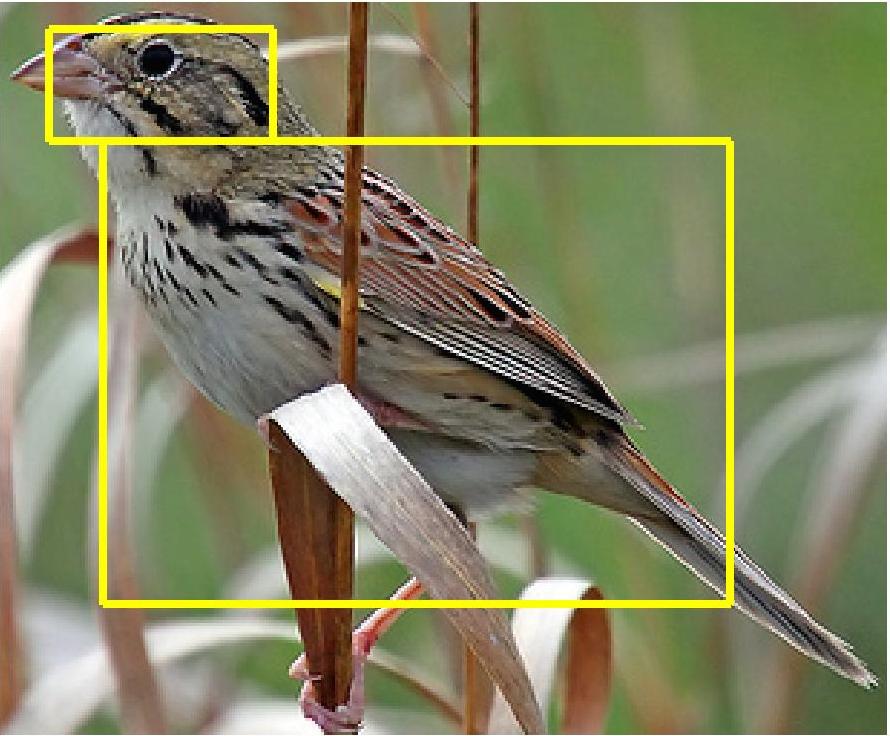} 
\includegraphics[width=0.15\linewidth, height = 0.1\linewidth]{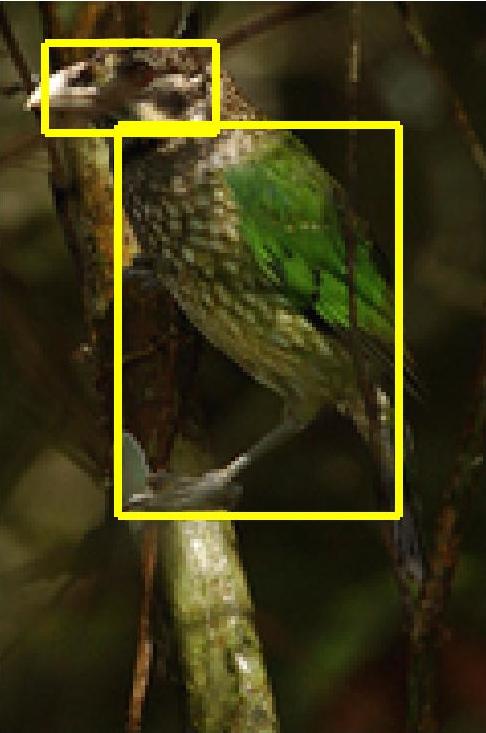} 
\includegraphics[width=0.15\linewidth, height = 0.1\linewidth]{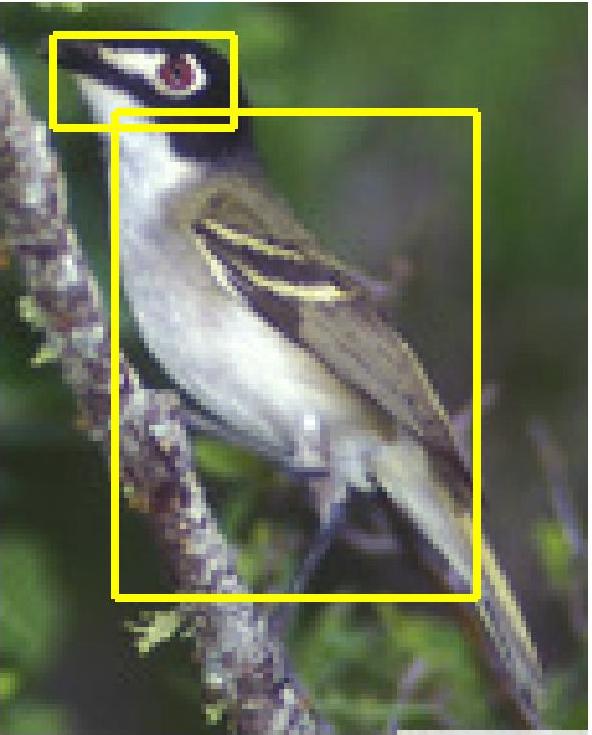} 
\includegraphics[width=0.15\linewidth, height = 0.1\linewidth]{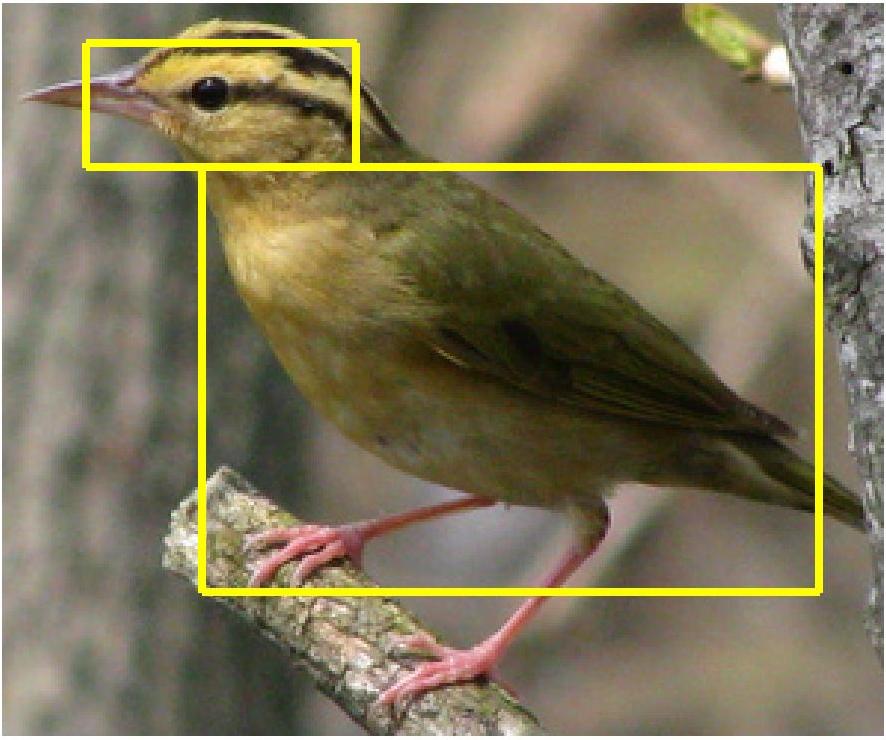} 
\end{center}
\caption{{Illustration of geometric constant $\delta^{NP}$. In each row, the first column is the test image with an R-CNN bounding box detection, and the rest are the top-five nearest neighbors in the training set, indexed using \texttt{pool5} features and cosine distance metric.}}
\label{fig:np}
\end{figure*}

\paragraph{Geometric constraints.}
Because the individual part detectors are less than perfect, the window with highest individual part detector scores is not always correct, especially when there are occlusions.
We therefore consider several scoring functions to enforce constraints over the layout of the parts relative to the object location to filter out incorrect detections.
We define
\begin{equation}
\Delta_{\mathrm{geometric}}(X) = \Delta_{\mathrm{box}}(X) \left( \prod_{i=1}^n \delta_i(x_i) \right)^{\alpha}
\end{equation}
where
$\delta_i$ is a scoring function for the position of the part $p_i$ given the training data. 
Following previous work on part localization from, e.g.~\cite{belkrieg,dpm,pictorial},
we experiment with three definitions of $\delta$:
\begin{itemize}

\item
$\delta_i^{MG}(x_i)$ fits a mixture of Gaussians model with $N_g$ components to the training data for part $p_i$.  In our experiments, we set $N_g = 4$.

\item
$\delta_i^{NP}(x_i)$ finds the $K$ nearest neighbors in appearance space to
$\tilde{x}_0$, where $\tilde{x}_0 = \arg \max d_0(x_0)$ is the top-scoring window from the root detector.
We then fit a Gaussian model to these $K$ neighbors.
In our experiments, we set $K = 20$. Figure \ref{fig:np} illustrates some examples of nearest neighbors. 
\end{itemize}

The DPM~\cite{dpm} models deformation costs with a per-component Gaussian prior. R-CNN~\cite{rcnn} is a single-component model, motivating the $\delta^{MG}$ or $\delta^{NP}$ definitions.
Our $\delta^{NP}$ definition is inspired by Belhumeur et al.~\cite{belkrieg}, but differs in that we index nearest neighbors on appearance rather than geometry.

%

\subsection{Fine-grained categorization}
We extract semantic features from localized parts as well as the whole object. The final feature representation is $[\phi(x_0) \ldots \phi(x_n)]$ where $x_0$ and $x_{1 \ldots n}$ are whole-object and part location predictions inferred using one of the models from the previous section and $\phi(x_i)$ is the feature representation of part $x_i$.

In one set of experiments, we extract deep convolutional features $\phi(x_i)$ from an ImageNet pre-trained CNN, similar to DeCAF~\cite{decaf}.
In order to make the deep CNN-derived features more discriminative for the target task of fine-grained bird classification, we also fine-tune the ImageNet pre-trained CNN for the 200-way bird classification task from ground truth bounding box crops of the original CUB images.
In particular, we replace the original 1000-way \texttt{fc8} classification layer with a new 200-way \texttt{fc8} layer with randomly initialized weights drawn from a Gaussian with $\mu = 0$ and $\sigma = 0.01$.
We set fine-tuning learning rates as proposed by R-CNN~\cite{rcnn}, initializing the global rate to a tenth of the initial ImageNet learning rate and dropping it by a factor of 10 throughout training, but with a learning rate in the new \texttt{fc8} layer of 10 times the global learning rate.
For the whole object bounding box and each of the part bounding boxes, we independently finetune the ImageNet pre-trained CNN for classification on ground truth crops of each region warped to the $227 \times 227$ network input size, always with 16 pixels on each edge of the input serving as context as in R-CNN~\cite{rcnn}.
At test time, we extract features for the predicted whole object or part region using the network fine-tuned for that particular whole object or part.

For training the classifier, we employ a one-versus-all linear SVM using the final feature representation. For a new test image, we apply the whole and part detectors with the geometric scoring function to get detected part locations and use the features for prediction.
If a particular part $i$ was not detected anywhere in the test image (due to all proposals falling below the part detector's threshold, set to achieve high recall), we set its features $\phi(x_i) = \mathbf{0}$ (zero vector).

\section{Evaluation}
In this section, we present a comparative performance evaluation of our proposed method.
Specifically, we conduct experiments on the widely-used fine-grained benchmark Caltech-UCSD birds dataset \cite{DatasetCUB200} (CUB200-2011).
The classification task is to discriminate among 200 species of birds, and is challenging for computer vision systems due to the high degree of similarity between categories.
It contains 11,788 images of 200 bird species. Each image is annotated with its bounding box and the image coordinates of fifteen keypoints: the beak, back, breast, belly, forehead, crown, left eye, left leg, left wing, right eye, right leg, right wing, tail, nape and throat. We train and test on the splits included with the dataset, which contain around 30 training samples for each species.
Following the protocol of \cite{dpd}, we use two semantic parts for the bird dataset: head and body.

We use the open-source package Caffe~\cite{Jia13caffe} to extract deep features and fine-tune our CNNs. For object and part detections, we use the Caffe reference model, which is almost identical to the model used by Krizhevsky et al. in \cite{krizhevsky}. We refer deep features from each layer as \texttt{conv}$n$, \texttt{pool}$n$, or \texttt{fc}$n$ for the $n$th layer of the CNN, which is the output of a convolutional,
pooling, or fully connected layer respectively.
We use \texttt{fc6} to train R-CNN object and part detectors as well as image representation for classification.
For $\delta^{NP}$, nearest neighbors are computed using \texttt{pool5}  and cosine distance metric.

\subsection{Fine-grained categorization}
We first present results on the standard fine-grained categorization task associated with the Caltech-UCSD birds dataset.
The first set of results in Table~\ref{tab:finegrainedres} are achieved in the setting where the ground truth bounding box for the entire bird is known at test time, as most state-of-art methods assume, making the categorization task somewhat easier. 
In this setting, our part-based method with the local non-parametric geometric constraint $\delta^{NP}$ works the best without fine-tuning, achieving 68.1\% classification accuracy without fine-tuning.
Fine-tuning improves this result by a large margin, to over 76\%.
We compare our results against three state-of-the-art baseline approaches with results assuming the ground truth bounding box at test time. We use deep convolutional features as the authors of \cite{decaf}, but they use a HOG-based DPM as their part localization method. The increase in performance is likely due to better part localization (see Table \ref{tab:partlocalres}). Oracle method uses the ground truth bounding box and part annotations for both training and test time. 

The second set of results is in the less artificial setting where the bird bounding box is \emph{unknown} at test time. Most of the literature on this dataset doesn't  report performance in this more difficult, but more realistic setting. As Table \ref{tab:finegrainedres} shows, in this setting our part-based method works much better than the baseline DPD model. We achieve 66.0\% classification accuracy without finetuning , almost as good as the accuracy we can achieve when the ground truth bounding box is given. This means there is no need to annotate any box during test time to classify the bird species. With finetuned CNN models, our method achieves 73.89\% classification accuracy. 
We are unaware of any other published results in this more difficult setting, but we note that our method outperforms previous state-of-the-art even without knowledge of the ground truth bounding box.

Another interesting experiment we did is to remove the part descriptors by only looking at the image descriptors inside the predicted bounding box. By having geometric constraints over part locations relative to object location, our method is able to help localize the object. As Table \ref{tab:finegrained_noparts} shows, our method outperforms a single object detector using R-CNN, which means the geometric constraints helps our method better localize the object window. The detection of strong DPM is not as accurate as our method, which explains the performance drop.
The ``oracle'' method uses the ground truth bounding box and achieves 57.94\% accuracy, which is still much lower than the method in Table \ref{tab:finegrainedres} of using both image descriptors inside object and parts.

\begin{table}[t]
\centering
\caption{Fine-grained categorization results on CUB200-2011 bird dataset. -ft means extracting deep features from finetuned CNN models using each semantic part. Oracle method uses the ground truth bounding box and part annotations for both training and test time. } 
\begin{tabular}{|l|r|}
\hline
\multicolumn{2}{|c|}{Bounding Box Given} \\
\hline
DPD~\cite{dpd} & 50.98\% \\
DPD+DeCAF feature ~\cite{decaf} & 64.96\% \\
POOF~\cite{poof} & 56.78\% \\
Symbiotic Segmentation~\cite{iccv13_symbiotic} & 59.40\% \\
Alignment~\cite{iccv13_alignment} & 62.70\%\\
\hline
Oracle & 72.83\% \\
Oracle-ft & 82.02\%\\
\hline
Ours ($\Delta_{\mathrm{box}}$) & 67.55\% \\
Ours ($\Delta_{\mathrm{geometric}}$ with $\delta^{MG}$) & 67.98\% \\
Ours ($\Delta_{\mathrm{geometric}}$ with $\delta^{NP}$) & 68.07\% \\
Ours-ft ($\Delta_{\mathrm{box}}$) & 75.34\% \\
Ours-ft ($\Delta_{\mathrm{geometric}}$ with $\delta^{MG}$) &  \textbf{76.37\%}\\
Ours-ft ($\Delta_{\mathrm{geometric}}$ with $\delta^{NP}$) & 76.34\%\\
\hline
\hline
\multicolumn{2}{|c|}{Bounding Box Unknown} \\
\hline
DPD+DeCAF~\cite{decaf} with no bounding box & 44.94\% \\
Ours ($\Delta_{\mathrm{null}}$) & 64.57\% \\
Ours ($\Delta_{\mathrm{box}}$)& 65.22\% \\
Ours ($\Delta_{\mathrm{geometric}}$ with $\delta^{MG}$) &65.98\% \\
Ours ($\Delta_{\mathrm{geometric}}$ with $\delta^{NP}$) & 65.96\% \\
Ours-ft ($\Delta_{\mathrm{box}}$)& 72.73\% \\
Ours-ft ($\Delta_{\mathrm{geometric}}$ with $\delta^{MG}$) & 72.95\% \\
Ours-ft ($\Delta_{\mathrm{geometric}}$ with $\delta^{NP}$) & \textbf{73.89\%} \\
\hline
\end{tabular}
\label{tab:finegrainedres}
\end{table}

\begin{table}[t]
\centering
\caption{Fine-grained categorization results on CUB200-2011 bird dataset with \emph{no parts}. We trained a linear SVM using deep features on all the methods. Therefore only the bounding box prediction is the factor of difference. -ft is the result of extracting deep features from fine-tuned CNN model on bounding box patches. } \label{tab:finegrained_noparts}
\begin{tabular}{|l|r|}
\hline
Oracle (ground truth bounding box) & 57.94\%\\
Oracle-ft & 68.29\% \\
\hline 
Strong DPM \cite{Hossein_ECCV12} & 38.02\% \\
R-CNN~\cite{rcnn} & 51.05\% \\
\hline \hline
Ours ($\Delta_{\mathrm{box}}$)  & 50.17\% \\
Ours ($\Delta_{\mathrm{geometric}}$ with $\delta^{MG}$) & 51.83\% \\
Ours ($\Delta_{\mathrm{geometric}}$ with $\delta^{NP}$) & 52.38\%\\
Ours-ft ($\Delta_{\mathrm{box}}$)  &  62.13\%\\
Ours-ft ($\Delta_{\mathrm{geometric}}$ with $\delta^{MG}$) & 62.06\% \\
Ours-ft ($\Delta_{\mathrm{geometric}}$ with $\delta^{NP}$) & \textbf{62.75\%} \\
\hline
\end{tabular}
\end{table}

\subsection{Part localization}
We now present results evaluating in isolation the ability of our system to accurately localize parts.
Our results in Table~\ref{tab:partlocalres} are given in terms of the Percentage of Correctly Localized Parts (PCP) metric. 
For the first set of results, the whole object bounding box is given and the task is simply to correctly localize the parts inside of this bounding box, with parts having $\ge 0.5$ overlap with ground truth counted as correct.

For the second set of results, the PCP metric is computed on top-ranked parts predictions using the objective function described in Sec. 3.2.
Note that in this more realistic setting we do not assume knowledge of the ground truth bounding box at test time -- despite this limitation, our system produces accurate part localizations.

\begin{table}[t]
\centering
\caption{Recall of region proposals produced by selective search methods on CUB200-2011 bird dataset. We use ground truth part annotations to compute the recall, as defined by the proportion of ground truth boxes for which there exists a region proposal with overlap at least 0.5, 0.6 and 0.7 respectively.}\label{tab:selective_search_recall}
\begin{tabular}{|c|c|c|c|}
\hline
Overlap & 0.50 & 0.60 & 0.70\\
\hline
Bounding box & 96.70\% & 97.68\% & 89.50\% \\
Head &  93.34\% & 73.87\%& 37.57\%\\
Body & 96.70\% & 85.97\%&54.68\%\\
\hline
\end{tabular}
\end{table}

\begin{table}[t]
\centering
\caption{Part localization accuracy in terms of PCP (Percentage of Correctly Localized Parts) on the CUB200-2011 bird dataset. There are two different settings: with given bounding box and without bounding box. } 
\label{tab:partlocalres}
\begin{tabular}{|l|r|r|}
\hline
\multicolumn{3}{|c|}{Bounding Box Given} \\
\hline
& \multicolumn{1}{|c|}{Head}
& \multicolumn{1}{|c|}{Body}
\\
\hline
Strong DPM~\cite{Hossein_ECCV12} & 43.49\% & 75.15\% \\
Ours ($\Delta_{\mathrm{box}}$)   & 61.40\% & 65.42\% \\
Ours ($\Delta_{\mathrm{geometric}}$ with $\delta^{MG}$)& 66.03\% & 76.62\% \\
Ours ($\Delta_{\mathrm{geometric}}$ with $\delta^{NP}$) & \textbf{68.19\%} & \textbf{79.82\%} \\
\hline
\multicolumn{3}{|c|}{Bounding Box Unknown} \\
\hline
& \multicolumn{1}{|c|}{Head}
& \multicolumn{1}{|c|}{Body}
\\
\hline
Strong DPM~\cite{Hossein_ECCV12} & 37.44\% & 47.08\% \\
Ours ($\Delta_{\mathrm{null}}$  ) &60.50\% &  64.43\% \\
Ours ($\Delta_{\mathrm{box}}$)  & 60.56\% & 65.31\% \\
Ours ($\Delta_{\mathrm{geometric}}$ with $\delta^{MG}$)& \textbf{61.94\%} & 70.16\% \\ 
Ours ($\Delta_{\mathrm{geometric}}$ with $\delta^{NP}$) & 61.42\% & \textbf{70.68\%} \\
\hline
\end{tabular}
\end{table}

\begin{figure*}[t]
\begin{center}
\includegraphics[width=0.45\linewidth]{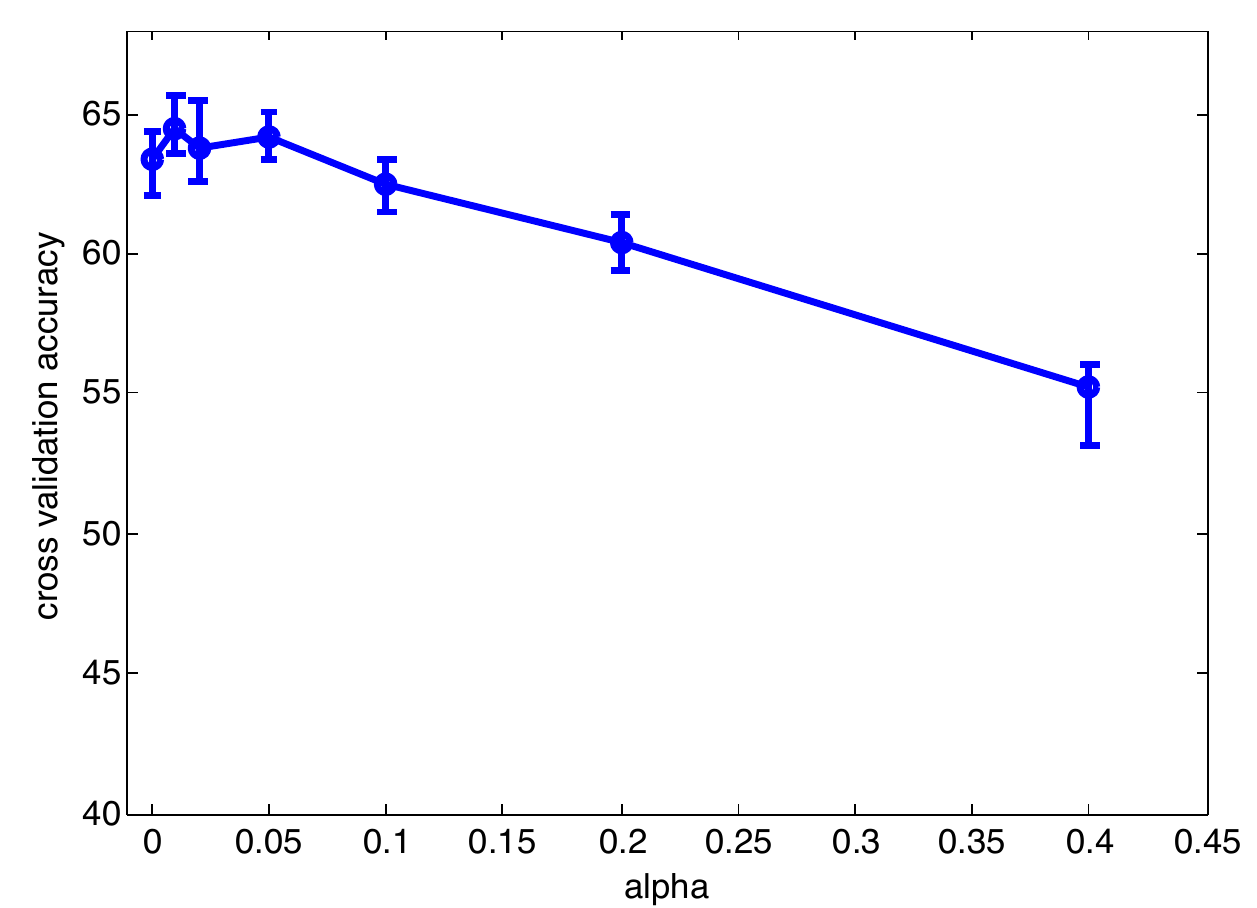}
\includegraphics[width=0.45\linewidth]{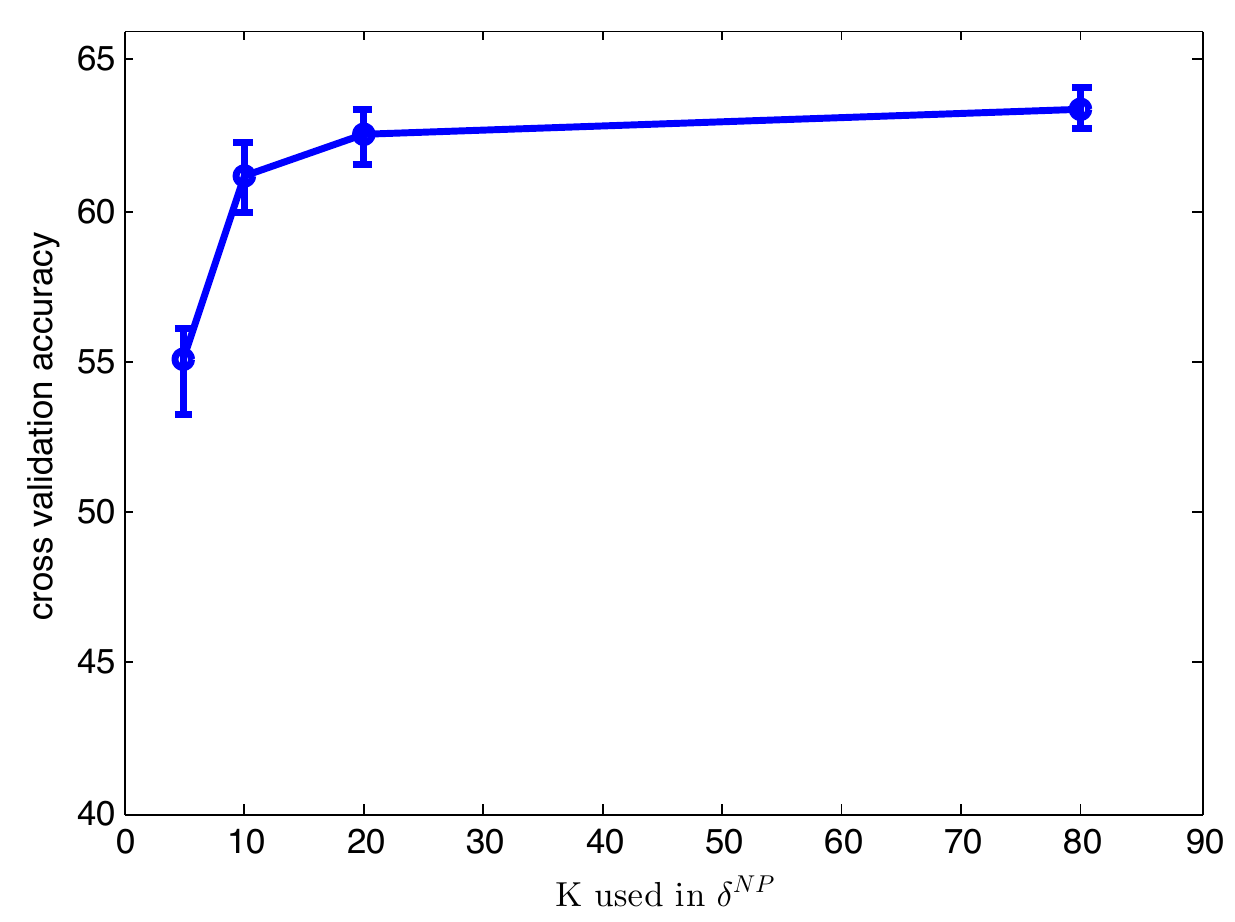}
\end{center}
\caption{Cross-validation results on fine-grained accuracy for different values of $\alpha$ (left) and $K$ (right). We split the training data into 5 folds and use cross-validate each hyperparameter setting.}
\label{fig:crossvalidationalphak}
\end{figure*}

As shown in Table \ref{tab:partlocalres}, for both settings of given bounding box and unknown bounding box, our methods outperform the strong DPM~\cite{Hossein_ECCV12} method.
Adding a geometric constraint $\delta^{NP}$ improves our results (79.82\% for body localization compared to 65.42\%). In the fully automatic setting, the top ranked detection and part localization performance on head is 65\% better than the baseline method. $\Delta_{\mathrm{null}}=1$ is the appearance-only case with no geometric constraints applied. Although the fine-grained classification results don't show a big gap between $\Delta_{\mathrm{geometric}}$ and $\Delta_{\mathrm{box}}$, we can see the performance gap for part localization.
The reason for the small performance gap might be that deep convolutional features are invariant to small translations and rotations,
limiting the impact of small localization errors on our end-to-end accuracy.

We also evaluate the recall performance of selective search region proposals \cite{selsearch} for bounding box and semantic parts. 
The results of recall given different overlapping thresholds are shown in Table \ref{tab:selective_search_recall}. 
Recall for the bird head and body parts is high when the overlap requirement is $0.5$, which provides the foundation for localizing these parts given the region proposals. However, we also observe that the recall for head is below $40\%$ when the overlap threshold is $0.7$, indicating the bottom-up region proposals could be a bottleneck for precise part localization.

\begin{figure*}
\begin{center}
\begin{tabular}{ccc}
\includegraphics[width=0.3\linewidth]{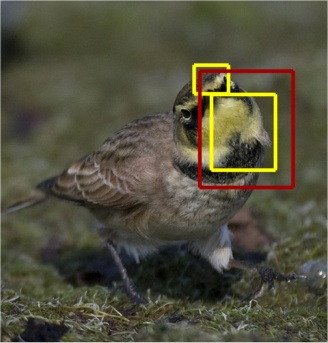} &
\includegraphics[width=0.3\linewidth]{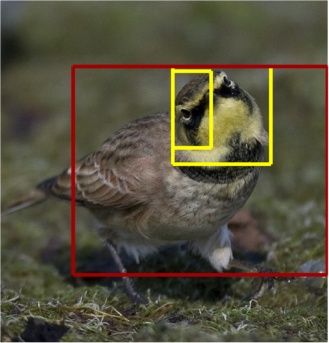} &
\includegraphics[width=0.3\linewidth]{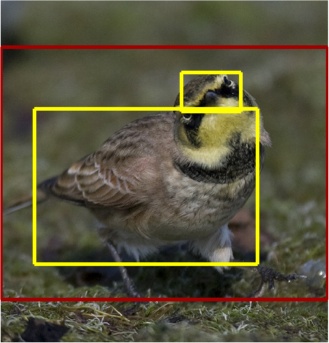} \\
\includegraphics[trim=0mm 10mm 0mm 10mm, clip, width=0.3\linewidth]{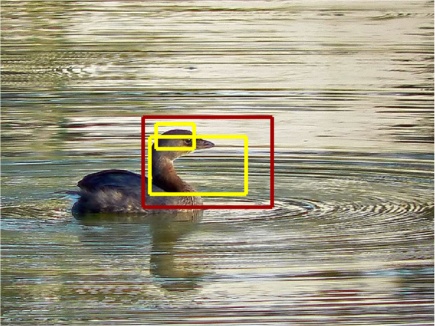} &
\includegraphics[trim=0mm 10mm 0mm 10mm, clip, width=0.3\linewidth]{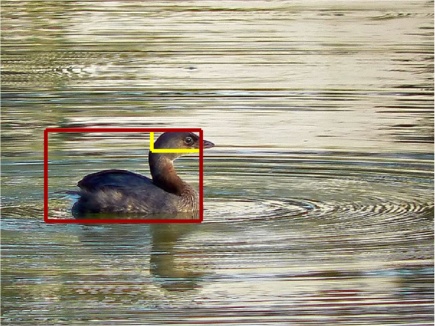} &
\includegraphics[trim=0mm 10mm 0mm 10mm, clip, width=0.3\linewidth]{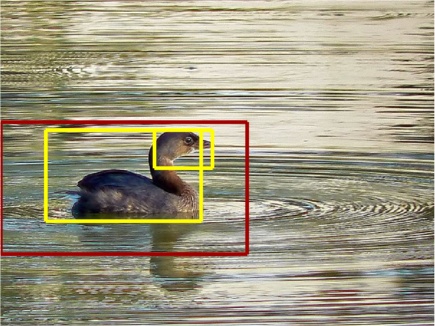} \\
\includegraphics[width=0.3\linewidth]{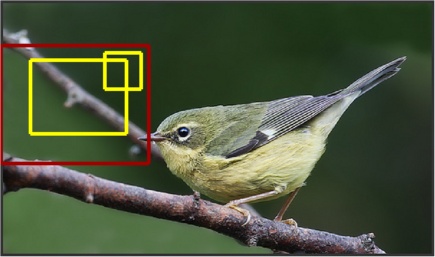} &
\includegraphics[width=0.3\linewidth]{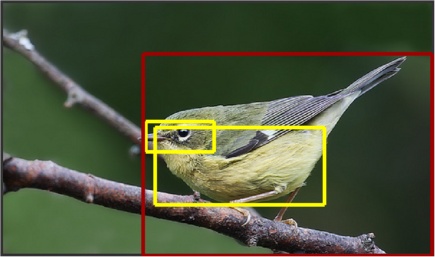} &
\includegraphics[width=0.3\linewidth]{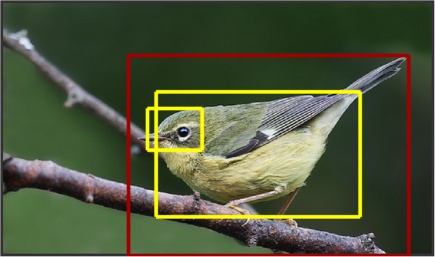} \\
\includegraphics[trim=0mm 20mm 0mm 20mm, clip, width=0.3\linewidth]{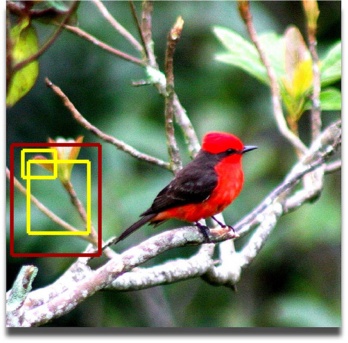} &
\includegraphics[trim=0mm 20mm 0mm 20mm, clip, width=0.3\linewidth]{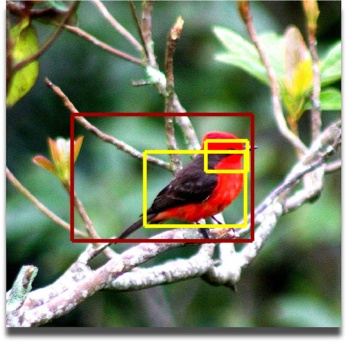} &
\includegraphics[trim=0mm 20mm 0mm 20mm, clip, width=0.3\linewidth]{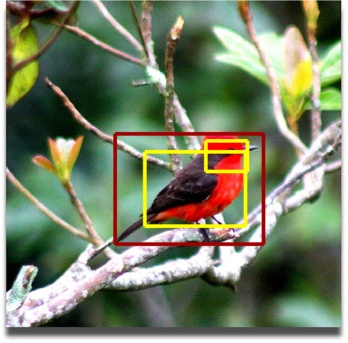} \\
\includegraphics[width=0.3\linewidth]{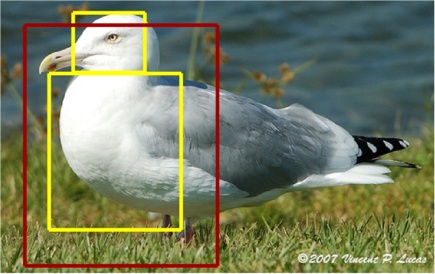} &
\includegraphics[width=0.3\linewidth]{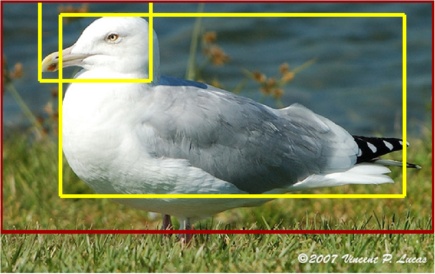} &
\includegraphics[width=0.3\linewidth]{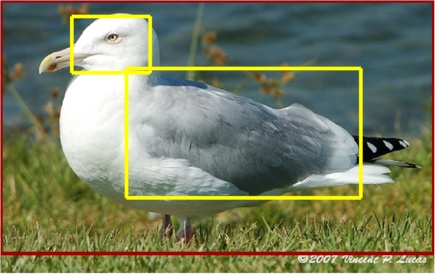} \\
\includegraphics[width=0.3\linewidth]{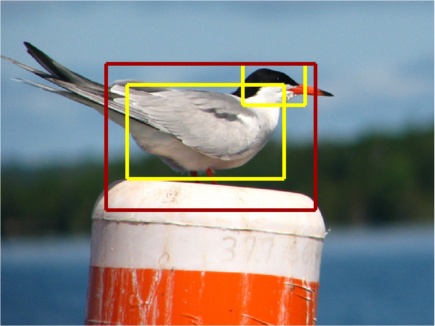} &
\includegraphics[width=0.3\linewidth]{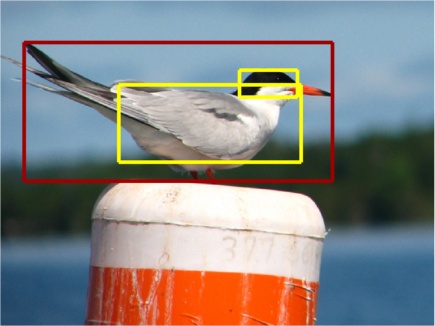} &
\includegraphics[width=0.3\linewidth]{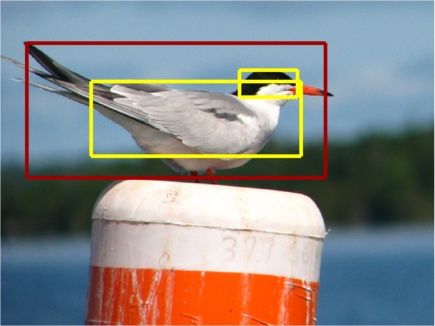} \\
Strong DPM & Ours ($\Delta_{box}$) & Ours ($\delta^{NP}$)
\\
\end{tabular}
\end{center}
\caption{{Examples of bird detection and part localization from strong DPM~\cite{Hossein_ECCV12} (left); our method using $\Delta_{\mathrm{box}}$ part predictions (middle); and our method using $\delta^{NP}$(right). All detection and localization results without any assumption of bounding box. }}
\label{fig:comparasion}
\end{figure*}

\begin{figure*}
\begin{center}
\includegraphics[height=0.2\linewidth]{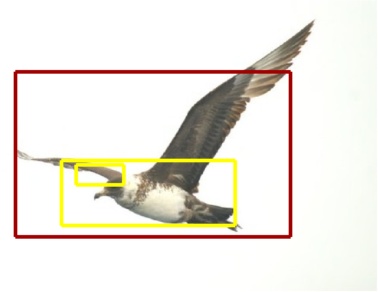} 
\includegraphics[height=0.2\linewidth]{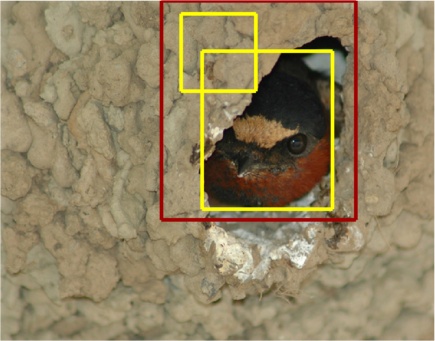} 
\includegraphics[height=0.2\linewidth]{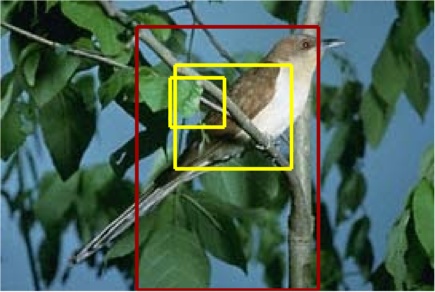} 
\includegraphics[height=0.2\linewidth]{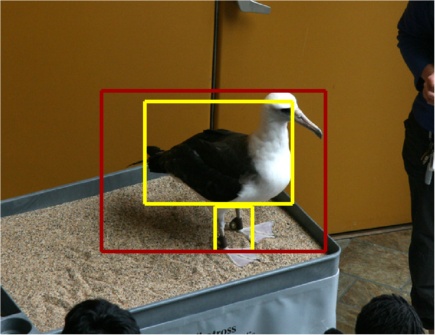} 
\includegraphics[height=0.2\linewidth]{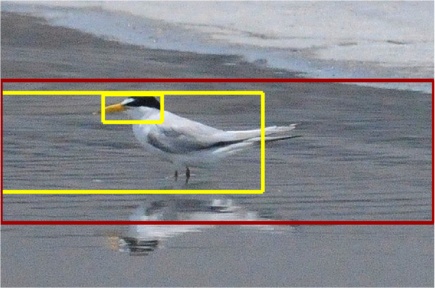}
\includegraphics[height=0.2\linewidth]{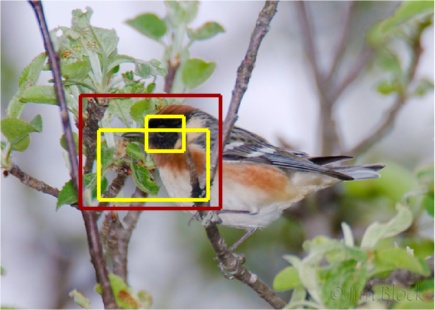} 
\includegraphics[height=0.2\linewidth]{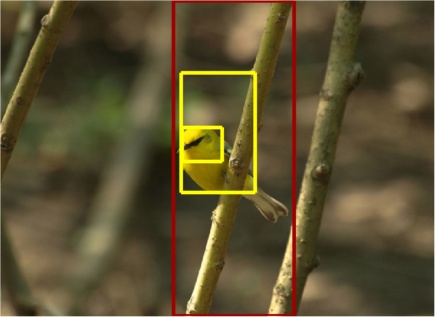} 
\includegraphics[height=0.2\linewidth]{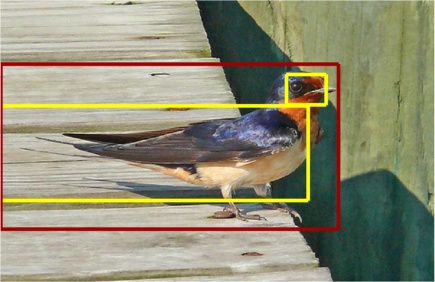} 
\includegraphics[height=0.2\linewidth]{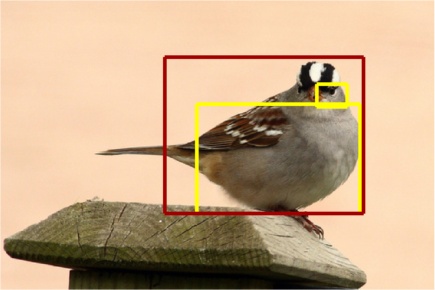} 
\end{center}
\caption{{Failure cases of our part localization using $\delta^{NP}$.}}
\label{fig:failure}
\end{figure*}

Other visualizations are shown in Figure~\ref{fig:comparasion}. We show three detection and part localization for each image, the first column is the output from strong DPM, the second column is our methods with individual part predictions and the last column is our method with local prior. We used the model pretrained from \cite{Hossein_ECCV12} to get the results. We also show some failure cases of our method in Figure~\ref{fig:failure}.

\subsection{Component Analysis}
To examine the effect of different values of $\alpha$ and $K$ used in $\Delta_{\mathrm{geometric}}$, we conduct cross-validation experiments.
Results are shown in Figure~\ref{fig:crossvalidationalphak}. We fix $K=20$ in Figure~\ref{fig:crossvalidationalphak}, left and fix $\alpha = 0.1$ in Figure \ref{fig:crossvalidationalphak}, right. All the experiments on conducted on training data in a cross-validation fashion and we split the training data into 5 folds.
As the results show, the end-to-end fine-grained classification results are sensitive to the choice of $\alpha$ and $\alpha=0$ is the case of $\Delta_{\mathrm{box}}$ predictions without any geometric constraints. The reason why we have to pick a small $\alpha$ is the pdf of the Gaussian is large compared to the logistic score function output from our part detectors. On the other hand, the choice of $K$ cannot be too small and it is not very sensitive when $K$ is larger than 10.

%

\section{Conclusion}
We have proposed a system for joint object detection and part localization capable of state-of-the-art fine-grained object recognition. Our method learns detectors and part models and enforces learned geometric constraints between parts and with the object frame. 
Our experimental results demonstrate that even with a very strong feature representation and object detection system, it is highly beneficial to additionally model an object's pose by means of parts for the difficult task of fine-grained discrimination between categories with high semantic similarity.
In future extensions of this work, we will consider methods which jointly model at training time the object category and each of its parts and deformation costs.
We also plan to explore the weakly supervised setting in which we automatically discover and model parts as latent variables from only the object bounding box annotations.
Finally, we will consider relaxing the use of selective search for smaller parts and employing dense window sampling.

\subsubsection{Acknowledgments} This work was supported in part by DARPA Mind's Eye and MSEE
programs, by NSF awards IIS-0905647, IIS-1134072, and IIS-1212798, and
by support from Toyota.

\newpage
\bibliographystyle{splncs03}
\bibliography{reference}
\end{document}